\documentclass[12pt]{spieman}  
\usepackage{amsmath,amsfonts,amssymb}
\usepackage{graphicx}
\usepackage{setspace}
\usepackage{tocloft}
\usepackage{authblk}
\usepackage{lineno}
\usepackage{multirow}
\usepackage{amsmath,scalerel}
\usepackage{bm}
\usepackage{subcaption}
\title{From Preoperative CT to Postmastoidectomy Mesh Construction: Mastoidectomy Shape Prediction for Cochlear Implant Surgery}

\author[a*]{Yike Zhang}
\author[b]{Eduardo Davalos}
\author[c]{Dingjie Su}
\author[d]{Ange Lou}
\author[c]{Jack Noble}

\affil[a]{St. Mary's University, San Antonio, TX, USA}
\affil[b]{Trinity University, San Antonio, TX, USA}
\affil[c]{Vanderbilt University, Nashville, TN, USA}
\affil[d]{Center for Advanced AI, Accenture, CA, USA}

\newcommand{\absdiv}[1]{%
  \par\addvspace{.5\baselineskip}
  \noindent\textbf{#1}\quad\ignorespaces
}
\cftpagenumbersoff{figure}
\cftpagenumbersoff{table} 
\begin{document} 
\maketitle

\begin{abstract}
\absdiv{Purpose:} Cochlear Implant (CI) surgery treats severe hearing loss by inserting an electrode array into the cochlea to stimulate the auditory nerve. An important step in this procedure is mastoidectomy, which removes part of the mastoid region of the temporal bone to provide surgical access. Accurate mastoidectomy shape prediction from preoperative imaging improves presurgical planning, reduces risks, and enhances surgical outcomes. Despite its importance, there are limited deep-learning-based studies regarding this topic due to the challenges of acquiring ground-truth labels. We address this gap by investigating self-supervised and weakly-supervised learning models to predict the mastoidectomy region without human annotations.

\absdiv{Approach:} We propose a hybrid self-supervised and weakly-supervised learning framework to predict the mastoidectomy region directly from preoperative CT scans, where the mastoid remains intact. Our self-supervised learning approach reconstructs the postmastoidectomy 3D surface from preoperative imaging, aim to align with the corresponding intraoperative microscope views for future surgical navigation related applications. Postoperative CT scans are used in the self-supervised learning model to assist training procedures despite additional challenges such as metal artifacts and low signal-to-noise ratios introduced by them. To further improve the accuracy and robustness, we introduce a Mamba-based weakly-supervised model that refines mastoidectomy shape prediction by using a novel 3D T-Distribution loss function, inspired by the Student-\textit{t} distribution. Weak supervision is achieved by leveraging segmentation results from the prior self-supervised framework, eliminating manual data labeling process.

\absdiv{Results:} Our hybrid method achieves a mean Dice score of 0.72 when predicting the complex and boundary-less mastoidectomy shape, surpassing state-of-the-art approaches and demonstrating strong performance. The method provides groundwork for constructing 3D postmastoidectomy surfaces directly from the corresponding preoperative CT scans.

\absdiv{Conclusion:} To our knowledge, this is the first work that integrating self-supervised and weakly-supervised learning for mastoidectomy shape prediction, offering a robust and efficient solution for CI surgical planning while leveraging 3D T-distribution loss in weakly-supervised medical imaging.
\end{abstract}


{\noindent \footnotesize\textbf{*}Yike Zhang,  \linkable{yzhang5@stmarytx.edu} }

\begin{spacing}{2}   

\section{Introduction}
\begin{figure}
    \centering
    \includegraphics[width=\textwidth]{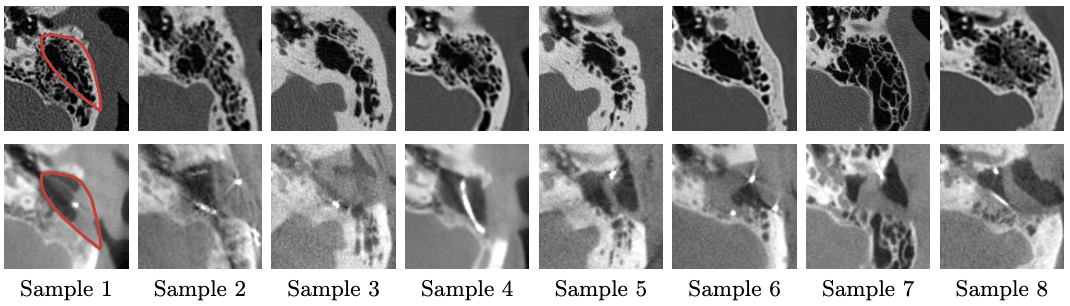}
    \caption{\textbf{Example CT scans and Mastoidectomy Region}. \textbf{Top row} shows preoperative CT scans and \textbf{bottom row} illustrates the corresponding highly noisy postoperative CT scans. The highlighted area in red is an example of the mastoidectomy shape region.}
    \label{fig:preop_postop}
\end{figure}

Cochlear Implant (CI) surgery is a widely used treatment for individuals with moderate-to-profound hearing loss, enabling auditory simulation using an electrode array that is inserted into the cochlea \cite{labadie2018preliminary}. A critical step in this procedure is mastoidectomy, where a portion of the temporal bone is removed using a high-speed drill to provide access to the cochlea for electrode placement. Accurate prediction of the mastoidectomy shape from preoperative Computed Tomography (CT) scans could play a crucial role in presurgical planning, robotic surgical assistance, and intraoperative navigation. Our research aims to directly predict the mastoidectomy region from preoperative CT scans. Postmastoidectomy CT scans are often acquired after the surgery to confirm electrode placement and assess surgical outcomes. However, they are unavailable of preoperative estimation of the surgical site, since only preoperative CT scans are available before. While these scans are used during training as noisy labels in our self-supervised framework, they are not used during inference. The reconstructed postmastoidectomy surface derived directly from preoperative CT scans could serve as an essential anatomical landmark for developing CI intraoperative navigation systems. However, the irregular and non-enclosed geometry of the mastoid region, composed of pneumatized bone with numerous small air cells, makes this task highly challenging for applying traditional segmentation approaches. Additionally, while postoperative CT scans capture the final mastoidectomy shape, they suffer from low resolution, intensity heterogeneity, and metal artifacts from implant components, limiting their usability in training deep-learning-based prediction models. Figure~\ref{fig:preop_postop} shows the final mastoidectomy cavity in postoperative CT scans. As seen in the bottom row of the figure, postoperative CT scans contain metal artifacts from implant electrode arrays and intensity heterogeneity due to fluid accumulation in the ear canal after surgery.
Previous research has explored various strategies for mastoidectomy shape prediction. One approach involves robot-assisted mastoidectomy, where a bone-attached robotic system performs the procedure with high precision \cite{Dillon2015A}. However, this method still requires surgeons to manually annotate the mastoidectomy region on preoperative CT scans, making it inefficient and time-consuming. Alternatively, image synthesis methods based on Generative Adversarial Networks (GANs) \cite{ciclegan} and diffusion models \cite{ddpm} have been proposed to generate postoperative-like images from preoperative CTs. Yet, these methods risk introducing artifacts such as false electrodes or wires, making them unsuitable for our research goals. Similarly, segmentation models like SAM and SAM2-derived medical imaging approaches \cite{SAM4MIS, lou2024zeroshotsurgicaltoolsegmentation} struggle with the ambiguous and complex boundaries of the mastoidectomy shape, leading to suboptimal performance. 

To address these challenges, we propose a hybrid self-supervised and weakly-supervised learning framework for automatic mastoidectomy shape prediction directly from preoperative CT scans. Our approach consists of two key components:
\begin{itemize}
    \item \textbf{Self-Supervised Learning using Postoperative Scans:} We first introduce a self-supervised neural network that predicts the mastoidectomy shape directly from preoperative CT images. By utilizing postoperative CT scans in the self-supervised learning framework, our method bypasses the need for labor-intensive manual annotations while ensuring segmentation accuracy and robustness. 
    \item \textbf{Weakly-Supervised Learning with 3D T-Distribution Loss:} To further improve the mastoidectomy shape prediction results, we employ a Mamba-based weakly-supervised learning model, leveraging segmentation results from the prior self-supervised framework as weak labels. We introduce a 3D T-Distribution loss function to effectively capture the irregularity and geometric variability of mastoidectomy regions while eliminating reliance on manually labeled training data.
\end{itemize}
In this paper, we use the Dice similarity coefficient (Dice score) as one of the key metrics for measuring segmentation accuracy. Dice score\cite{Dice1945-gh} is a statistical measure of overlap between two samples, commonly used to evaluate segmentation accuracy, with values ranging from 0 to 1. Our proposed hybrid framework achieves a mean Dice score of 0.72, surpassing state-of-the-art models such as UNetr \cite{unetr} and SwinUNetr \cite{SwinUnetr}. To the best of our knowledge, this is the first work to combine self-supervised and weakly-supervised learning for mastoidectomy shape prediction, providing an efficient and clinically relevant solution for CI surgical planning and robot-assisted procedures. Furthermore, our findings demonstrate the broader applicability of 3D T-Distribution loss in weakly-supervised medical imaging, providing novel approaches for segmentation tasks involving complex anatomical structures.

The contributions of our paper can be summarized as follows:
\begin{itemize}
    \item \textbf{First Comprehensive Dataset for Mastoidectomy Shape Prediction:} We construct a comprehensive mastoidectomy prediction dataset using raw preoperative and postoperative CT scans, overcoming challenges such as metal artifacts, low signal-to-noise ratio, and intensity heterogeneity. To the best of our knowledge, this is the first dataset specifically designed for the mastoidectomy shape prediction, enabling future CI research advancements in robot-assisted surgery, preoperative surface visualization, and surgical tool tracking.
    \item \textbf{Hybrid Self-Supervised and Weakly-Supervised Learning for Mastoidectomy Shape Prediction:} We propose a novel hybrid deep-learning-based framework for segmenting the mastoidectomy region from the preoperative CT scans and ultimately reconstructing the 3D postmastoidectomy surface. By leveraging preoperative/postoperative CT scans and weak labels, our approach eliminates the need for manual annotations, improving efficiency in the training pipeline.
    \item \textbf{Introduction of 3D T-Distribution Loss for Weakly-Supervised Medical Imaging:} We introduce a 3D T-Distribution loss function, which effectively captures the geometric variability and irregular boundaries of mastoidectomy regions. Our results demonstrate that 3D T-Distribution loss improves prediction accuracy in weakly-supervised learning, making it a robust loss function for medical imaging tasks involving noisy and complex anatomical structures.
\end{itemize}
\section{Method}
\subsection{Self-supervised Learning Framework}\label{sec:ssl_method}
\begin{figure}[!ht]
  \centering
  \includegraphics[width=\linewidth]{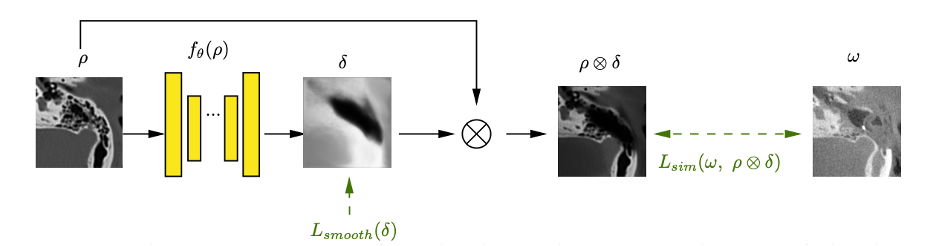}
  \caption{Framework overview. We compare the predicted mastoidectomy region applied to the preoperative CT $\rho \otimes \delta$ against the postoperative CT $\omega$ using similarity-based loss function $L_{msssim\_cscc}$. A smoothing term $L_{smooth}$ is also applied to $\delta$ to further reduce noise output by network $f_{\theta}$.}
  \label{fig:ssl_overview}
\end{figure}

\begin{figure}[!ht]
  \centering
  \includegraphics[width=\linewidth]{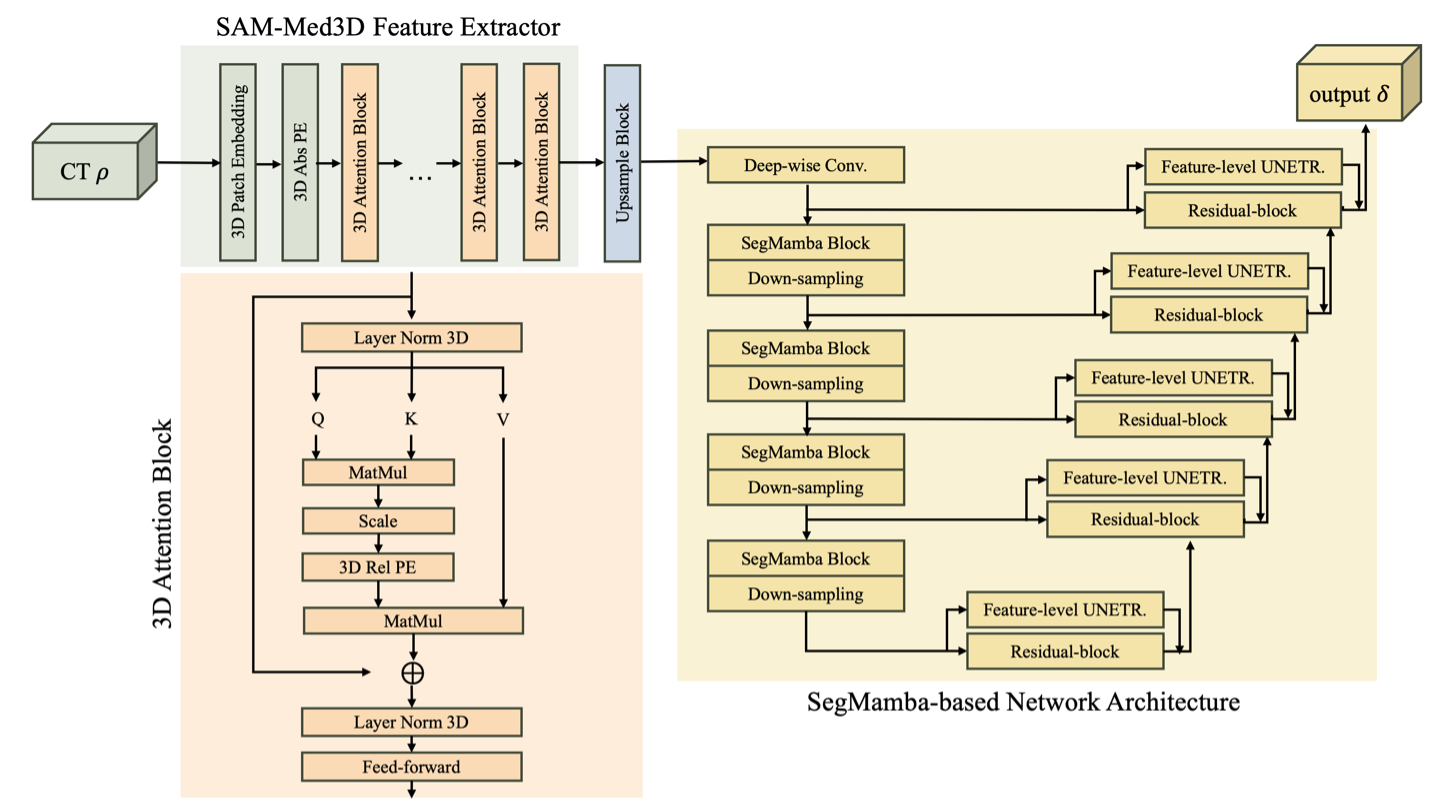}
  \caption{SegMamba-based Neural Network with pretrained SAM-Med3D Image Encoder}
  \label{fig:ssl_network_structure}
\end{figure}

In this work, we used a dataset of 751 preoperative and postoperative CT pairs from patients who underwent cochlear implantation. The data were collected at Vanderbilt University Medical Center, using CT scanners from multiple vendors. We randomly assigned 630 cases for training and validation, while reserving the rest for testing. We manually annotated the mastoidectomy region in 32 randomly selected samples from the test dataset for evaluation. As preprocessing before training our method, the preoperative CT $\rho$ and postoperative CT $\omega$ were aligned via rigid registration, and their intensity values were normalized. 
As shown in Fig. \ref{fig:ssl_overview}, the removed mastoid volume in $\omega$ tends to have lower intensities compared to surrounding tissue since it only contains air and soft tissue. Thus, we develop a model to predict the mastoidectomy region (represented as an inverted probability map $\delta$) on $\rho$ and make the predicted post-mastoid CT ($\rho \otimes \delta$) have higher similarity when compared with $\omega$. The neural network shown in Fig. \ref{fig:ssl_network_structure} is modeled as a function $f_{\theta}$($\rho$) = $\delta$ using the state-of-the-art SegMamba-based \cite{xing2024segmambalongrangesequentialmodeling} neural network with a pretrained SAM-Med3D \cite{wang2023sammed3d} encoder as our feature extractor, where $\theta$ is the set of neural network parameters. In our model, we reduce the original CT dimensions to $[160, 160, 64]$ by cropping around the ear regions using an ear anatomy segmentation neural network proposed in Zhang et al.\cite{zhang2023} . 

The Multi-Scale Structural Similarity Index (MS-SSIM) \cite{msssim} represents an improvement over the traditional Structural Similarity Index (SSIM) \cite{ssim}. This enhancement enables MS-SSIM to comprehensively evaluate the similarities and differences between two images across multiple scales. It effectively accommodates the wide variety of heterogeneity that can occur between two different images. We extend this metric with Squared Cross-Correlation (SCC) \cite{scc} to quantify the overall structural similarity and distribution consistency between two image volumes $\rho \otimes \delta$ and $\omega$. We propose two self-supervised loss functions $L_{msssim\_cscc}$ and $L_{smooth}$ to train our model $f_{\theta}$. $L_{msssim\_cscc}$ effectively minimizes the impact of undesired noise present in postoperative CTs. However, this loss function may result in fragmentation within the output probability map, thus we add the second loss function $L_{smooth}$ to address the potential edge sharpness and encourage smoother value transitions in the output volume. The $L_{smooth}$ is defined as $L_{\text{smooth}}(\delta) = \sum_{i=1}^{N}\left\| \nabla \left( \delta_{i} \right) \right\|^2$. We define $\frac{\partial{\delta}}{\partial{x}} \approx \delta(i_x+1, i_y, i_z) - \delta(i_x, i_y, i_z)$, and the same rule applies to the $\frac{\partial{\delta}}{\partial{y}}$ and $\frac{\partial{\delta}}{\partial{z}}$. $L_{msssim\_cscc}$ performs across various scales, employing a multi-step down-sampling technique. Applying $L_{msssim\_cscc}$ alone effectively captures the target mastoidectomy region even when $\omega$ is contaminated by various noises. 
\begin{multline}
L_{\text{msssim\_cscc}}(\rho \otimes \delta, \omega) = 1 - [l_M(\rho \otimes \delta, \omega)]^{\alpha_M}
\cdot \prod_{j=1}^{M} [c_j(\rho \otimes \delta, \omega) + SCC_j(\rho \otimes \delta, \omega)]^{\beta_j} [s_j(\rho \otimes \delta, \omega)]^{\gamma_j}
\label{Eq:L_msssim_cscc}
\end{multline} In Eq. \ref{Eq:L_msssim_cscc}, $M$ represents the total number of scales at which the comparison is performed, and we assign $M$=5 in the loss function\cite{msssim}. $\alpha_M$ is the weight given to the luminance component at the coarsest scale. $\beta_j$ refers to the weights assigned to the contrast components at each scale $j$. $\gamma_j$ are the weights for the structure components at each scale $j$. $l$($\rho \otimes \delta$, $\omega$) is the luminance comparison function that measures the similarity in brightness between two image volumes. $c$($\rho \otimes \delta$, $\omega$) is the contrast comparison function that evaluates the similarity in contrast between two image volumes. $s$($\rho \otimes \delta$, $\omega$) is the structure comparison function that represents the similarity in structure or pattern between two image volumes. 
SCC is being added to the contrast comparison function in this equation to enhance MS-SSIM capability to further maximize the distribution similarity at each scaling level. SCC is defined as:
\begin{equation}
    \text{SCC}(
\rho \otimes \delta, \omega) = \frac{\Bigl(\sum_{i=1}^{N}\left [\left( (\rho \otimes \delta)_{i} - \overline{
\rho \otimes \delta} \right) \left( \omega_{i} - \overline{\omega} \right) \right ]\Bigl)^2}{\sum_{i=1}^{N}\left(  (\rho \otimes \delta)_{i} - \overline{
\rho \otimes \delta}  \right)^2 \sum_{i=1}^{N}\left(  \omega_{i} - \overline{\omega}  \right)^2},
\label{Eq:scc}
\end{equation} where $N$ is noted as the total number of pixels.
\subsection{Weakly-supervised Learning Framework}
\begin{figure}[htb]
\centering
\begin{minipage}[b]{\linewidth}
  \includegraphics[width=\textwidth]{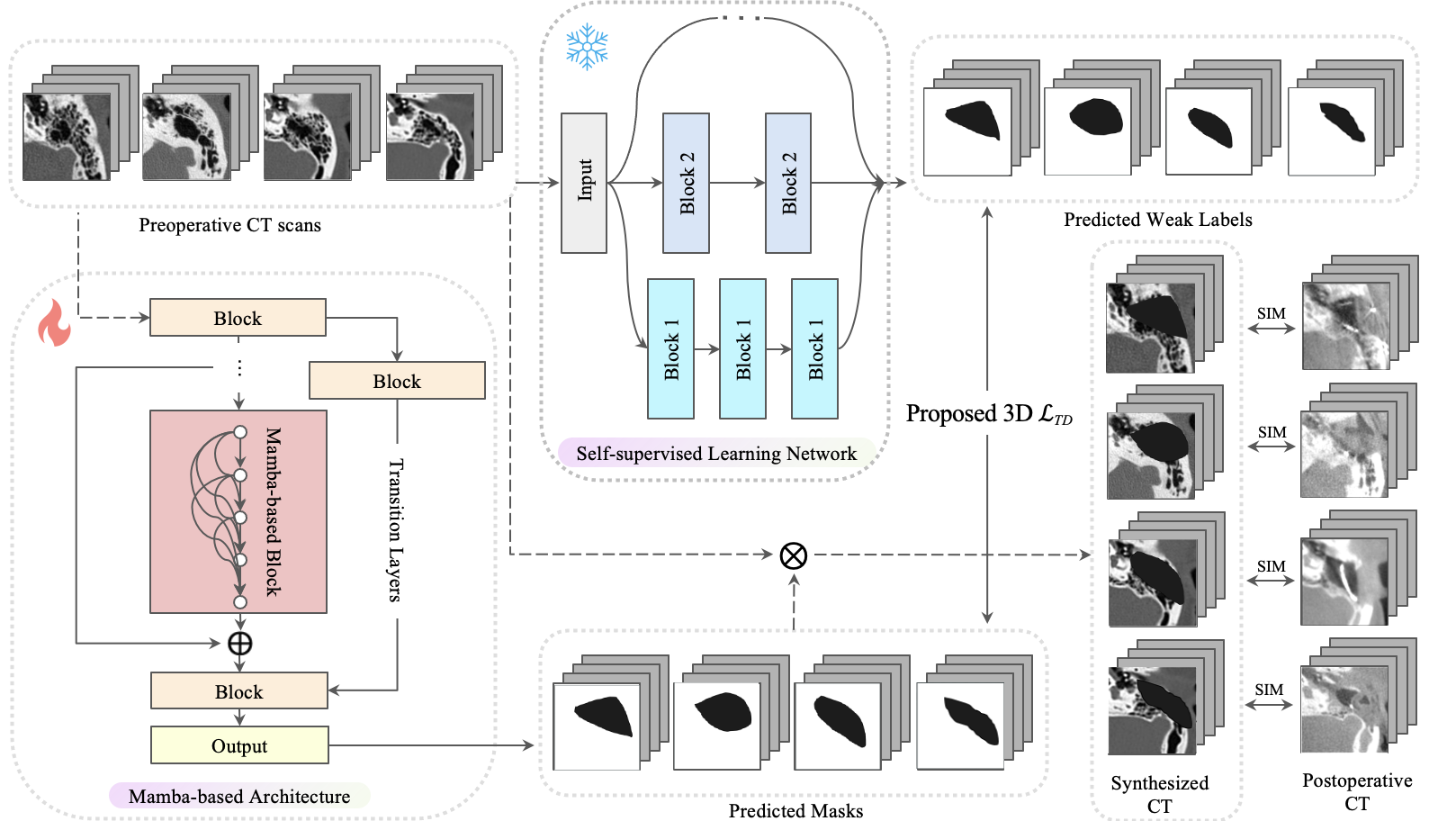}
  \caption{\textbf{Framework Overview}. \textcolor{cyan}{Snowflake} indicate frozen and non-trainable parameters, \textcolor{red}{fire} represents trainable parameters, and ``SIM" compares the similarity between two objects.}
\label{fig:wl_overview}
\end{minipage}
\end{figure}

Figure~\ref{fig:wl_overview} demonstrates the training and inference pipeline in our proposed framework. The preoperative scans are represented as $\bm{I} \in \mathbb{R}^{D \times H \times W}$, where $D$, $H$, and $W$ correspond to the depth, height, and width dimensions of the CT scans. In our dataset, $D=64$, $H=160$, and $W=160$. These volumetric images are then input into the SegMamba-based neural network $f_{\bm{\theta}}(\bm{I})$, which outputs probability masks $\bm{Y} \in \mathbb{R}^{D \times H \times W}$, representing the predicted mastoidectomy shape. During the weakly-supervised training process, we refine these predictions by comparing $\bm{Y}$ with weak labels $\bm{K} \in \{0, 1\}^{D \times H \times W}$, which are generated according to the self-supervised Mamba-based segmentation method \cite{zhang2024mmunsupervisedmambabasedmastoidectomy}. To further improve the accuracy of the predicted masks and mitigate the noise and large errors introduced by the weak labels, we propose the 3D T-Distribution loss, $\mathcal{L}_{TD}(\bm{Y}, \bm{K})$, which refines the probability masks and aligns them more closely with the actual mastoidectomy shape region.

\begin{figure}[htb]
\begin{minipage}[b]{1.0\linewidth}
  \centering
  \centerline{\includegraphics[width=\textwidth]{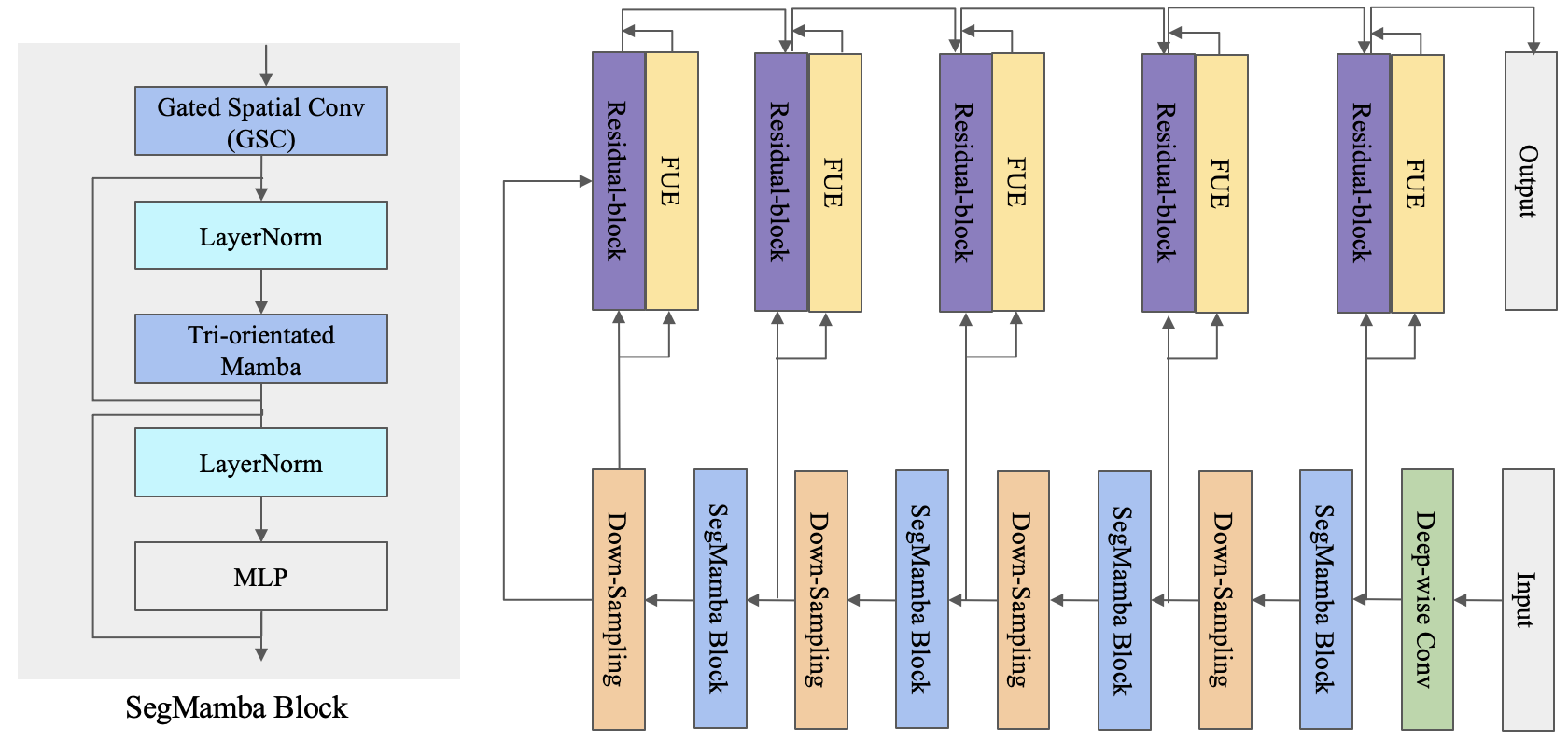}}
\caption{\textbf{Mamba-based Architecture}. The model takes preoperative CT as input and outputs the predicted mastoidectomy shape.}
\label{fig:segmamba}
\end{minipage}
\end{figure}

In this study, we use the state-of-the-art SegMamba \cite{xing2024segmambalongrangesequentialmodeling} neural network architecture shown in Figure~\ref{fig:segmamba} to train the mastoidectomy shape prediction network $f_{\bm{\theta}}$. This is a novel framework that combines the U-Net \cite{unet} shape-like structure with the newly released Mamba \cite{gu2024mambalineartimesequencemodeling} for modeling the global features in 3D image volumes at various scales.
Mamba is developed based on state space models and aims to capture long-range dependencies and enhance the efficiency of training and inference. SegMamba architecture contains a gated spatial convolution (GSC) module, the tri-orientated Mamba (ToM) module, and a feature-level uncertainty estimation (FUE) module. GSC module improves the spatial feature representation before each ToM module, ToM module supports whole-volume sequential modeling of 3D features, and FUE module selects and reuses the multi-scale features from the encoder.

The 3D T-Distribution loss function extends the 2D-imaging-focused method proposed by \cite{gonzalezjimenezRobustTLoss2023} and is specifically tailored for 3D imaging segmentation tasks. T-Distribution loss is derived from the negative log-likelihood of the Student-\textit{t} distribution, which is known for its ability to model data with heavy tails. The Probability Density Function (PDF) of the multivariate continuous Student-\textit{t} distribution is defined in Eq~\ref{eq:pdf}. 
\begin{equation}
    P(\bm{K}|\bm{\mu},\bm{\Sigma};r)=
    (\pi r)^{-D'/2} \; \cdot \frac{\Gamma(\frac{r+D'}{2})}{\Gamma(\frac{r}{2})} \; \cdot |\det(\bm{\Sigma})|^{-1/2} \; \cdot \\
    \left[ 1 + \frac{(\bm{K}-\bm{\mu})^T \bm{\Sigma}^{-1}(\bm{K}-\bm{\mu})}{r} \right]^{-\frac{r+D'}{2}},
\label{eq:pdf}
\end{equation} where $r$ represents the distribution's degree of freedom, $D'$ is the dimensional variable, $\bm{\mu}$ is the mean, and $\bm{\Sigma}$ is the covariance matrix of the distribution. Different from the Gaussian distribution, which is sensitive to outliers, the Student-\textit{t} distribution has a broader bell shape, making it more robust when encountering data points that deviate largely from the norm. The robustness in T-Distribution loss is achieved by controlling the variable $r$. A smaller $r$ creates a heavier tail thus reducing the influence of outliers by giving less weight to large errors. In contrast, a larger $r$ makes the distribution more Gaussian, increasing the sensitivity to outliers. Given this attribute, T-Distribution loss can be tuned to work with imperfect datasets more effectively than traditional loss functions. This makes it particularly valuable in medical image processing, where outliers and noise are common due to artifacts or variations in imaging quality. Furthermore, when applying T-Distribution loss to the weakly-supervised learning task where noisy labels are used during training, it helps improve the accuracy of predictions by mitigating the impact of outliers, leading to robust and reliable outcomes. The negative log-likelihood loss term can be represented in Eq~\ref{Eq: tloss}:
\begin{multline}
        \mathcal{L}_{TD} = \frac{D'}{2}\log(\pi r) + \log \Gamma \left(\frac{r}{2}\right) - \log \Gamma \left(\frac{r + D'}{2}\right) + \\ + \frac{1}{2}\sum_{i=1}^{D'}\log\sigma_{i}^{2} + \frac{r+D'}{2} \log \left[1 + \frac{1}{r}\sum_{i=1}^{D'}\frac{(K_i - u_i)^2}{\sigma_{i}^{2}}\right]
\label{Eq: tloss}
\end{multline} We set $\delta$ = $\bm{K} - f_{\bm{\theta}}(\bm{I})$ and $\sigma^{2}$ are the diagonal elements of the covariance matrix $\bm{\Sigma}$. Since $\bm{\Sigma}$ is symmetric and positive semi-definite, it is possible to only calculate the diagonal and lower triangular elements because the upper triangular part is simply mirroring of the lower triangular part. However, the dimensional data in our dataset is large, thus optimizing the full covariance matrix becomes computationally expensive and it is not feasible to be implemented under our current experiment settings.
In our experiment, we improve the overall performance by configuring the covariance matrix, $\bm{\Sigma}$, as a diagonal matrix rather than an identity matrix. By updating the parameter $r$ and $\sigma^{2}$ alongside the network parameters $\bm{\theta}$ during backpropagation, the network dynamically adapts to varying levels and distributions of weak labels without prior knowledge. Since the network's predictions may include negative values, we apply the SoftPlus activation function to $\log\sigma_{i}^{2}$ to ensure the output values remain positive.
\section{Results}
\subsection{Self-supervised Learning Results}
\begin{figure}[!ht]
  \centering
  \includegraphics[width=0.9\linewidth]{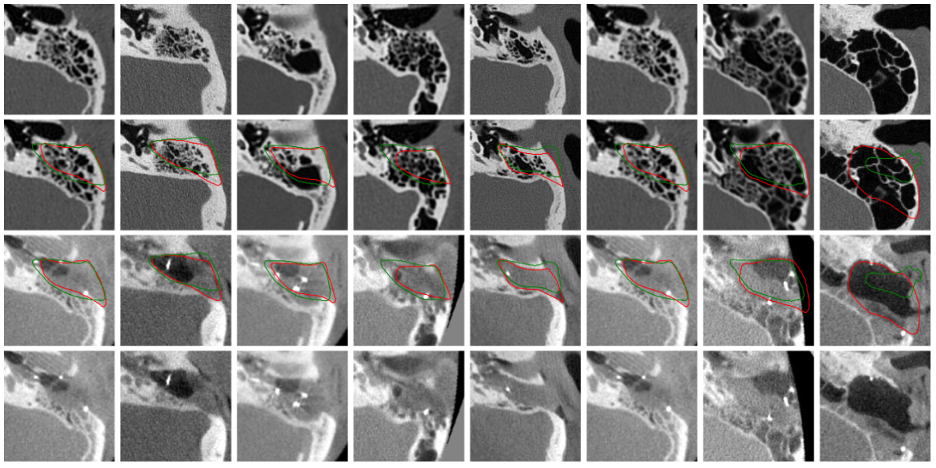}
\caption{Qualitative Performance Evaluation. The first row displays preoperative CT scans. The second row overlays predicted mastoidectomy areas (green) and ground truth (red) on preoperative CTs. The third row shows these contours on postoperative CT scans, and the final row displays the postoperative CTs. }
\label{fig:ssl_qualitative_output}
\end{figure}
Our dataset comprises a total of 751 images, divided into 504 training samples, 126 validation samples, and 121 testing samples. As stated in Section \ref{sec:ssl_method}, we manually annotated 32 ground truth mastoidectomy volume labels from the testing dataset. This annotation process, requiring approximately one hour per case, was highly labor-intensive due to the inherent heterogeneity between preoperative and postoperative CT scans, as well as the substantial variability in the shape and size of the removed mastoid regions. To assess the proposed model's performance in predicting mastoidectomy shapes within preoperative CT scans, we input voxel values from the 3D preoperative images into the Mamba-based model and binarize the resulting probability masks to generate the final predictions. Fig. \ref{fig:ssl_qualitative_output} shows the qualitative performance on eight test samples, with each column representing a test case. Our method predicts the mastoidectomy region well in the majority of the cases. The last column shows the case with the worst Dice score using our proposed method. In this case, the mastoid region in that preoperative CT scan is unusually sparse compared to the other samples in our test dataset. This leads to a predicted mastoidectomy region that is smaller than the ground truth using the self-supervised method.

\begin{figure}[!ht]
  \centering
  \includegraphics[width=0.6\textwidth]{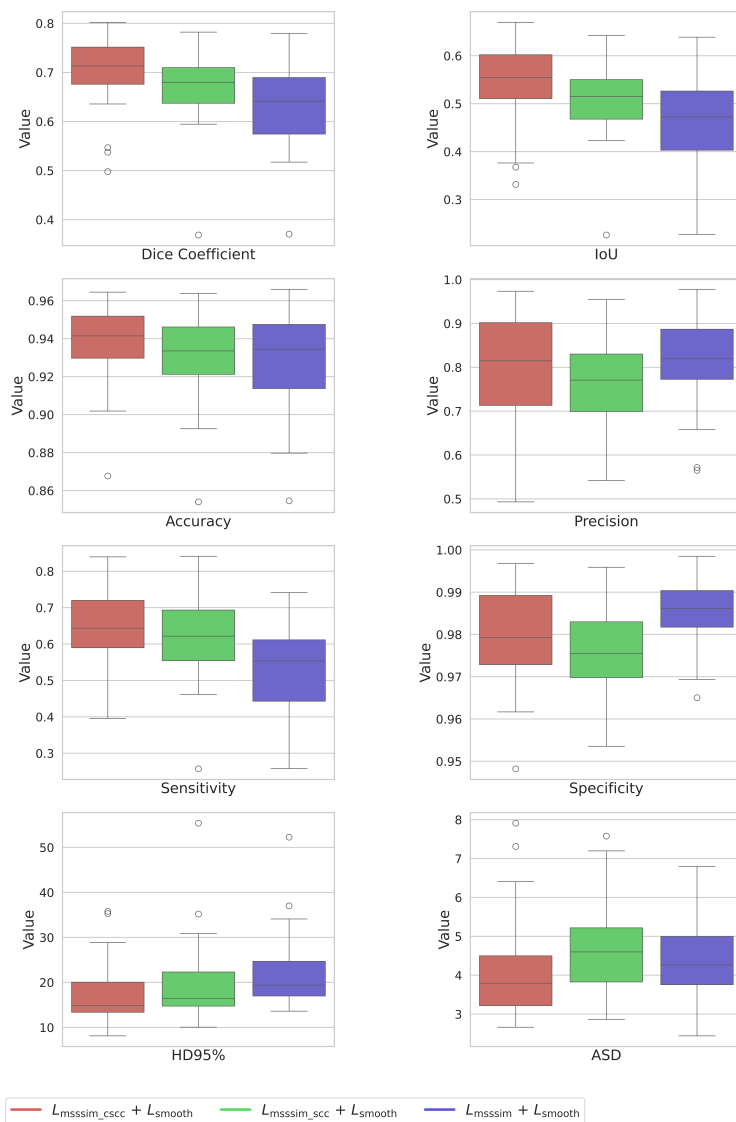}
\caption{Ablation Study: comparing performance across different loss functions.}
\label{fig:ablation_study}
\end{figure}

Additionally, Sørensen–Dice coefficient (Dice), Intersection over Union (IoU), Accuracy (Acc), Precision (Pre), Sensitivity (Sen), Specificity (Spe), Hausdorff Distance (HD95\%), and Average Surface Distance (ASD) are used to measure the similarities between the predicted mastoidectomy region by our neural network and ground truth labels in Table \ref{Tab:analysis}. The HD95\% metric is a modification of the Hausdorff Distance that focuses on the 95th percentile of the distance to reduce the influence of outliers. As shown in the table, it compares the aforementioned metrics of widely-used transformer-based networks and U-Net-based networks, including SwinUNetr \cite{SwinUnetr}, UNETR \cite{unetr}, the U-Net++ \cite{zhou2019unetplusplus} / U-Net \cite{unet} network using ImageNet \cite{imagenet} pretrained weights, and our proposed mamba-based method. The asterisk $^*$ followed by a number indicates that the difference between the proposed and competing methods is statistically significant based on a Wilcoxon test with a \textit{p}-value of less than 0.05. We observe that the predictions obtained by the proposed method tend to have lower sensitivity than specificity and precision. This suggests that our model is more likely to produce false negative values than false positive values, implying that the synthesized regions are generally smaller than the ground truth labels. Fig. \ref{fig:ablation_study} shows our method's ablation study with different loss functions. $L_{msssim\_cscc}$ is our proposed loss term, while $L_{msssim\_scc}$ refers to adding Eq. \ref{Eq:scc} to the s($\rho \otimes \delta$, $\omega$) structural similarity function instead of the c($\rho \otimes \delta$, $\omega$) contrast comparison function in MS-SSIM (Eq. \ref{Eq:L_msssim_cscc}). $L_{msssim}$ indicates the original MS-SSIM loss function. As shown in the figure, the proposed loss term outperforms the other combinations across the majority of metrics. 
\begin{table*}[ht]
    \centering
    \begin{center}
        \begin{tabular}{ l|c|c|c|c|c|c|c|c }
            \hline
            \multicolumn{1}{c|}{Methods} &
            \multicolumn{1}{c|}{Dice $\uparrow$} & 
            \multicolumn{1}{c|}{IoU $\uparrow$} & 
            \multicolumn{1}{c|}{Acc $\uparrow$} & 
            \multicolumn{1}{c|}{Pre $\uparrow$} & 
            \multicolumn{1}{c|}{Sen $\uparrow$} &
            \multicolumn{1}{c|}{Spe $\uparrow$} & 
            \multicolumn{1}{c|}{HD95\% $\downarrow$} &
            \multicolumn{1}{c}{ASD $\downarrow$} \\
            \hline
            UNet\cite{unet} & 0.5432$^{*}$ & 0.3867 & 0.9203 & 0.8215 & 0.4319 & 0.9877 & 26.1439 & 5.0094$^{*}$\\
            \hline
            UNet++\cite{zhou2019unetplusplus} & 0.6527$^{*}$ & 0.4931 & 0.9230 & 0.7937 & 0.6010 & 0.9671 & 20.3570 & 6.2326$^{*}$\\
            \hline
            UNetr\cite{unetr} & 0.5608$^{*}$ & 0.4020 & 0.9263 & \textbf{0.8734} & 0.4246 & \textbf{0.9936} & 20.1116 & 6.5478$^{*}$\\
            \hline
            SwinUNetr\cite{SwinUnetr} & 0.6426$^{*}$ & 0.4834 & 0.9317 & 0.8506 & 0.5293 & 0.9885 & \textbf{17.2194} & 5.6297$^{*}$\\
            \hline
            Our Method  & \textbf{0.7019} & \textbf{0.5450} & \textbf{0.9375} & 0.7970 & \textbf{0.6444} & 0.9795 & 17.3515 & \textbf{4.1585}\\
            \hline
        \end{tabular}
    \end{center}
    \caption{Performance comparison between Transformer/U-Net-based network and Mamba-based network}
    \label{Tab:analysis}
\end{table*}
\begin{figure}[!ht]
  \centering
  \includegraphics[width=0.8\textwidth]{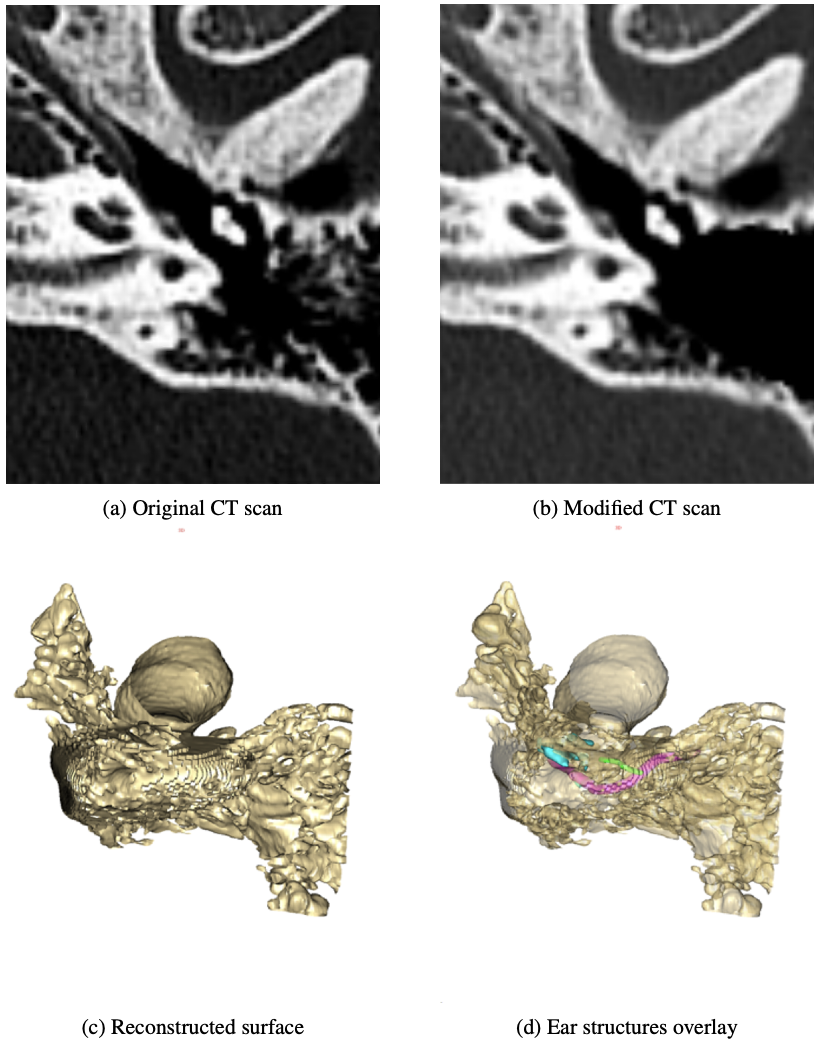}
\caption{Reconstructed post-mastoidectomy surface from the modified preoperative CT scan with the predicted mastoidectomy region. In (c) and (d), the gold surface represents the reconstructed post-mastoidectomy surface using our proposed method. In (d), Cyan highlights the ossicles, Lime highlights the chorda, and Magenta highlights the facial nerve. The ossicles (Cyan), chorda (Lime), and facial nerve (Magenta) are not reconstructed by our method but are instead extracted directly from the original CT scan.}
\label{fig:extracted_meshes}
\end{figure}
After the successful prediction of the mastoidectomy shape region using our proposed self-supervised learning method, we provide an additional example of post-mastoidectomy CT surface reconstruction in Fig. \ref{fig:extracted_meshes}, achieved by applying isosurfacing to the preoperative CT scan modified with the predicted mastoidectomy shape. Specifically, Fig. \ref{fig:extracted_meshes}a shows the original preoperative CT scan. Fig. \ref{fig:extracted_meshes}b illustrates the preoperative CT scan masked with the predicted mastoidectomy region. Fig. \ref{fig:extracted_meshes}c displays the corresponding CT mesh reconstructed using the isosurface function applied to the modified preoperative CT scan. In Fig. \ref{fig:extracted_meshes}d, the reconstructed surface is shown with an overlay of essential ear structures, including the ossicles, facial nerve, and chorda, providing a comprehensive visualization of the overall post-mastoidectomy CT reconstruction results.
\subsection{Weakly-supervised Learning Results}
To increase variations in the training data, we perform the following data augmentation techniques: random swapping, flipping, elastic deformations, and affine transformations. The initial learning rates for $\theta$, $r$, and $\sigma^2$, are set to $1 \times 10^{-3}$, $1 \times 10^{-4}$, and $1 \times 10^{-4}$, respectively. The 3D T-Distribution loss is initialized with $r$ = 1 and $\sigma^{2}$ as an identity matrix. A safeguard number $1e-8$ is applied to both of them for numerical stability. All models are trained for up to 600 epochs on an NVIDIA RTX 4090 GPU with an early stopping technique based on $\mathcal{L}_{TD}$. The overall training duration is approximately 24 to 48 hours.

The Dice similarity coefficient (Dice) and Hausdorff Distance 95\% (HD95) are used as part of the evaluation metrics for experiment results. HD95 calculates the 95th percentile of surface distances between ground truth and prediction point sets. Metric formulations are presented in the following:
\begin{equation}
    \text{Dice} = \frac{2\Sigma_{i=1}^N Y_i \bar{Y}_i}{\Sigma_{i=1}^N Y_i + \Sigma_{i=1}^N  \bar{Y_i}},
\end{equation} where N denotes the number of pixels. $Y$ and $\bar{Y}$ represent the output probability and ground truth masks, respectively.
\begin{equation}
    \text{HD95} = \max \{\underset{y^\prime \in Y^\prime}{\max} \underset{\bar{y^\prime}\in \bar{Y}^{\prime}}{\min} || y^\prime - \bar{y}^\prime ||_{0.95}, \underset{\bar{y}^\prime \in Y^\prime}{\max} \underset{y^\prime \in \bar{Y}^{\prime}}{\min} || \bar{y}^\prime - y^\prime||_{0.95} \},
\end{equation} where $y^\prime$ and $\bar{Y}^{\prime}$ denote ground truth and prediction surface point sets. Average Surface Distance (ASD) is also used in the evaluation. Unlike volume-based overlap metrics, such as Dice Score and Specificity, the HD95 and ASD evaluate the agreement between the surfaces of the ground truth labels and predicted structures, with a fixed tolerance. This offers a more precise evaluation of how closely those two surfaces align.
\begin{figure}[!ht]
  \centering
  \includegraphics[width=\linewidth]{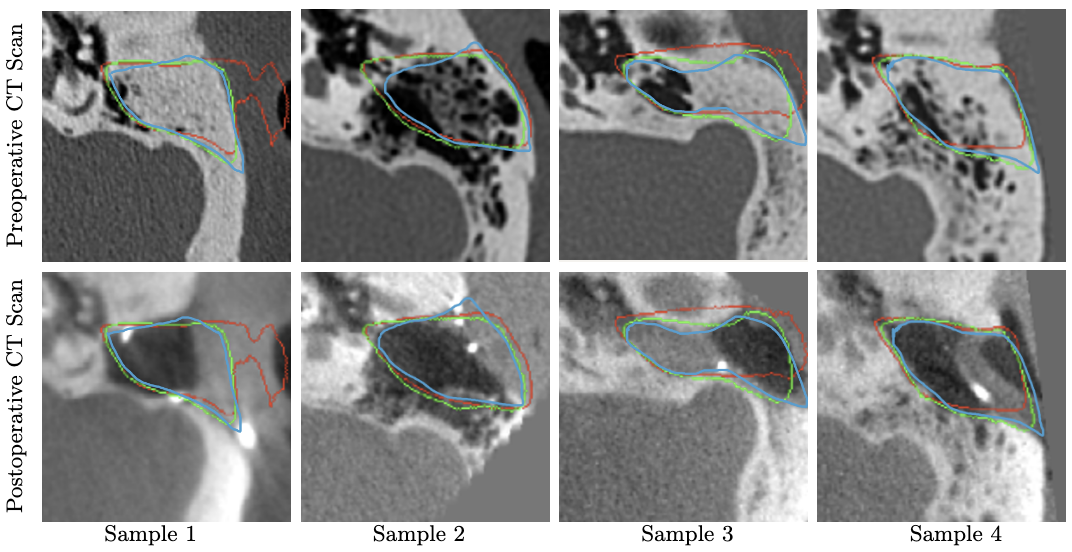}
\caption{\textbf{Mastoidectomy Shape Prediction}. Ground truth labels are highlighted in blue, predictions from the proposed method in green, and the baseline method from \cite{zhang2024mmunsupervisedmambabasedmastoidectomy} in red.}
\label{fig:ws_output}
\end{figure}
\begin{figure}[!ht]
  \centering
  \includegraphics[width=\linewidth]{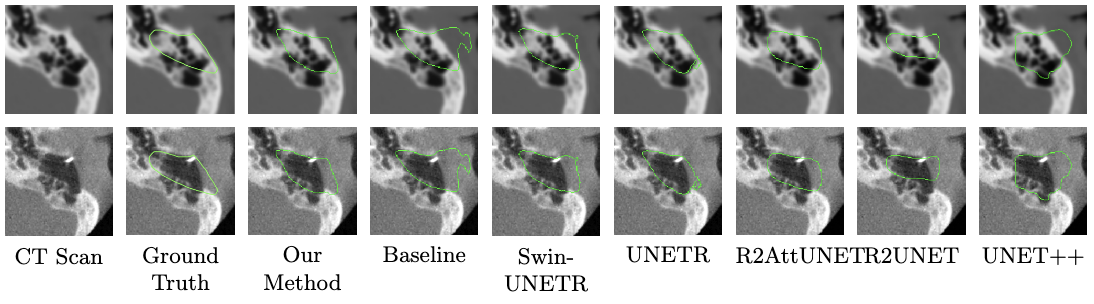}
\caption{\textbf{Representative Samples}. Qualitative visualizations of predicted mastoidectomy regions that are highlighted in green using different methods. The \textbf{first row} shows mastoidectomy shape predictions (in green) on preoperative images, while the \textbf{second row} shows predictions on postoperative images.}
\label{fig:comparing_samples}
\end{figure}
\begin{table*}[ht]
\centering
\begin{tabular}{l | lccccccc}
\hline
Methods & Dice$\uparrow$ & IoU$\uparrow$ & Acc$\uparrow$ & Pre$\uparrow$ & Sen$\uparrow$ & Spe$\uparrow$ & HD95$\downarrow$ & ASD$\downarrow$ \\
\hline
UNET++ & 0.328$^{*}$ & 0.203 & 0.889 & 0.563 & 0.244 & 0.977 & 32.367 & 9.349$^*$ \\
R2UNET & 0.484$^{*}$ & 0.325 & 0.914 & 0.848 & 0.348 & 0.992 & 27.948 & 5.273$^*$ \\
R2AttUNET & 0.497$^{*}$ & 0.343 & 0.917 & \textbf{0.891} & 0.361 & \textbf{0.995} & 24.377 & 5.435$^*$ \\
UNETR & 0.698$^{*}$ & 0.542 & 0.931 & 0.821 & 0.624 & 0.983 & 17.612 & 3.890$^*$ \\
Baseline & 0.702  & 0.545 & 0.938 & 0.797 & 0.644 & 0.980 & 17.352 & 4.159$^{*}$ \\
SwinUNETR & 0.712 & 0.557 & 0.939 & 0.806 & 0.655 & 0.981 & 16.930 & 3.971$^{*}$ \\
Proposed Method & \textbf{0.721} & \textbf{0.568} & \textbf{0.941} & 0.826 & \textbf{0.656} & 0.983 & \textbf{16.159} & \textbf{3.638} \\
\hline
\end{tabular}
\caption{\textbf{Quantitative Evaluation}. The proposed method achieves state-of-the-art performance in most metrics compared to widely-used U-Net-based and transformer-based networks. Note: IoU, Acc, Pre, Sen, Spe are the abbreviations of Intersection over Union, Accuracy, Precision, Sensitivity, and Specification, respectively.}
\label{tab:table1}
\end{table*} 
Our dataset consists of 751 images with 504 training, 126 validation, and 121 testing samples. In the test set, we manually annotate the ground truth mastoidectomy volume labels for 32 random testing cases. The annotation process, which took approximately 1 hour per case, was labor-intensive due to the heterogeneity between preoperative and postoperative CT scans and the significant variability in the shape and size of the removed mastoid regions. To evaluate the model's ability to segment the mastoidectomy shape within preoperative CT scans, we begin by querying the Mamba-based model with voxel values from the 3D preoperative images and binarizing the predicted probability masks. The results are then used to compare with outputs from the baseline method \cite{zhang2024mmunsupervisedmambabasedmastoidectomy} and the annotated ground truth labels. The results of four mastoidectomy shape predictions are shown in Figure~\ref{fig:ws_output}.
Further representative samples are shown in Figure~\ref{fig:comparing_samples} to demonstrate the promising results of identifying target region by the proposed method when comparing with other state-of-the-art models, such as baseline \cite{zhang2024mmunsupervisedmambabasedmastoidectomy}, SwinUNETR \cite{SwinUnetr}, UNETR \cite{unetr}, Recurrent Residual U-Net (R2UNET) \cite{alom2018recurrentresidualconvolutionalneural}, Recurrent Residual Attention U-Net (R2AttUNET) \cite{r2aunet}, and the vanilla UNET++ \cite{zhou2018unetnestedunetarchitecture}. Note that all comparing models are trained with only $\mathcal{L}_{TD}$ and the best performance checkpoint is used for the evaluation.
Quantitative evaluation is presented in Table~\ref{tab:table1}. The results show that the proposed method outperforms others in most metrics. We obtain the state-of-the-art Dice score of \textbf{0.721} and achieve the lowest HD95 of \textbf{16.159}.
The asterisk ($^*$) followed by a number indicates that the difference between the proposed and competing methods is statistically significant, as determined by a Wilcoxon test with a \textit{p}-value $\leq$ 0.05 for the corresponding metric. High precision and specificity indicate that the predicted areas are generally smaller than the ground truth labels, leading to fewer false positives. Sensitivity, on the other hand, measures the model's ability to minimize false negatives, and higher sensitivity means fewer false negatives.
To assess the effectiveness of the proposed 3D T-Distribution Loss in handling noisy labels compared to other commonly used loss functions, we present an ablation study in Table~\ref{tab:table2}. The comparing loss functions include the Cross-Entropy Loss ($\mathcal{L}_{CE}$)\cite{mao2023crossentropylossfunctionstheoretical}, the Binary Cross-Entropy Loss ($\mathcal{L}_{BCE}$)\cite{8943952}, the Focal Loss ($\mathcal{L}_{FL}$)\cite{lin2018focallossdenseobject}, the Mean Squared Error Loss ($\mathcal{L}_{MSE}$)\cite{terven2025survey}, and the Mean Absolute Error Loss ($\mathcal{L}_{MAE}$)\cite{terven2025survey}. We evaluate the proposed 3D T-Distribution loss against commonly used loss functions under identical training configurations to ensure a fair comparison.
\begin{table}[t]
    \centering
    \begin{tabular}{p{2cm}cc}
        \hline
        \multirow{2}{*}{\textbf{Loss Function}} & \multicolumn{2}{c}{\textbf{Average Accuracy}} \\
        & Dice$\uparrow$ & HD95$\downarrow$ \\
        \hline
        $\mathcal{L}_{CE}$ & 0.667 $\pm$ 0.082 & 31.163 $\pm$ 8.214 \\
        $\mathcal{L}_{BCE}$ & 0.707 $\pm$ 0.069 & 17.780 $\pm$ 8.057\\
        $\mathcal{L}_{FL}$ & 0.707 $\pm$ 0.066 & 17.124 $\pm$ 7.246 \\
        $\mathcal{L}_{MSE}$ & 0.716 $\pm$ 0.064 & 16.932 $\pm$ 7.812 \\
        $\mathcal{L}_{MAE}$ & 0.719 $\pm$ 0.064 & 16.616 $\pm$ 7.301\\
        $\mathcal{L}_{TD}$ & \textbf{0.721 $\pm$ 0.066} & \textbf{16.159 $\pm$ 6.905} \\
        \hline
    \end{tabular}
    \caption{\textbf{Ablation Study}. Comparing the Dice and HD95 metrics with various widely-used loss functions. The T-Distribution loss function achieved the highest Dice of 0.721 and the lowest HD95 of 16.159.}
    \label{tab:table2}
\end{table}
As shown in Table~\ref{tab:table2}, $\mathcal{L}_{TD}$ is more robust to handle noisy labels and achieves the highest Dice score and the lowest HD95 value.
\section{Discussion and Conclusion}
In this paper, we propose a hybrid self-supervised and weakly-supervised learning framework for mastoidectomy shape prediction in Cochlear Implant (CI) surgery, eliminating the need for labor-intensive manual annotation. Our method achieves a mean Dice coefficient of 0.72, surpassing state-of-the-art approaches, and demonstrating superior accuracy and robustness. Within the self-supervised learning framework, we show that the Mamba-based architecture outperforms conventional U-Net and transformer-based networks. Furthermore, our weakly-supervised approach highlights the effectiveness of the proposed 3D T-distribution loss, which show greater robustness against noisy or weak labels compared to traditional loss functions.

While our framework demonstrates strong performance, several limitations should also be acknowledged. First, the dataset was collected from Vanderbilt University Medical Center and a small number of collaborating hospitals. Although CT scans from multiple vendors were included to increase variability, the model’s generalizability to broader clinical settings remains to be further validated. Second, our evaluation focused on mastoidectomy shape prediction, and applying the exact same framework to other anatomical bony structures may present unique challenges. Finally, the reconstructed mesh surfaces currently lack realistic textures, which may limit their immediate use in intraoperative navigation without further improvement.

Our findings highlight the advantages of applying self-supervised and weakly-supervised learning framework in medical imaging, particularly for procedures where manual labeling is time-consuming and impractical. From our quantitative and qualitative evaluations, the 3D T-Distribution loss function handles noise and outliers effectively, offering potential applicability for other 3D medical imaging segmentation tasks. Additionally, reconstructing the post-mastoidectomy surface enables various downstream applications, including intraoperative navigation, 3D surgical scene understanding, and in-depth surgical analysis.

For future work, we aim to apply realistic textures to the reconstructed mastoidectomy surface, improving its resemblance to the corresponding intraoperative views and generating synthetic multi-views for CI surgery. Other approaches that automatically generate weak labels, such as leveraging pre-trained models or atlas-based segmentation methods, could also eliminate the need for manual labeling process across a wider range of 3D medical imaging segmentation tasks, assisting preoperative planning and intraoperative navigation.

\section*{Disclosures}
The authors declare there are no financial interests, commercial affiliations, or other potential conflicts of interest that have influenced the objectivity of this research or the writing of this paper.

\section*{Code and Data Availability}
The data supporting the findings of this study is not publicly available due to the policy of the IRB under which the data was collected; however, the source scripts and pseudocode are available upon request.

\section*{Compliance with ethical standards}
\label{sec:ethics}
This study was performed in accordance with the principles of the Declaration of Helsinki. Approval was granted by the IRB of Vanderbilt University Medical Center. (IRB number: 090155).

\section*{Acknowledgments}
This work was supported in part by grants R01DC014037, R01DC008408, and R01DC022099 from the NIDCD. This work is solely the responsibility of the authors and does not necessarily reflect the views of this institute.

\bibliography{report}
\bibliographystyle{spiejour}


\vspace{2ex}\noindent\textbf{Yike Zhang} received her Ph.D. degree in Computer Science at Vanderbilt University, specializing in medical image analysis, deep learning, and image-guided surgery. Her research focuses on developing AI-driven surgical planning and intraoperative navigation systems, particularly for cochlear implant surgery. She has implemented self-supervised and weakly-supervised learning frameworks to address complex challenges in medical imaging. She is currently a tenure-track assistant professor in Software Engineering at St.Mary's University.
\listoffigures
\listoftables

\end{spacing}
\end{document}